\documentclass[sigplan]{acmart}

\AtBeginDocument{%
  \providecommand\BibTeX{{%
    \normalfont B\kern-0.5em{\scshape i\kern-0.25em b}\kern-0.8em\TeX}}}

\setcopyright{acmcopyright}
\copyrightyear{2022}
\acmYear{2022}
\acmDOI{nnnnnnn.nnnnnnn}

\usepackage{acronym}
\usepackage{array}
\newcolumntype{x}[1]{>{\centering\let\newline\\\arraybackslash\hspace{0pt}}p{#1}}

\begin{document}
\settopmatter{printacmref=false} 
\renewcommand\footnotetextcopyrightpermission[1]{} 
\pagestyle{plain} 

\title{Optimizing LLVM Pass Sequences with Shackleton: \\A Linear Genetic Programming Framework}

\author{Hannah Peeler}
\email{hannah.peeler@arm.com}
\affiliation{%
  \institution{Arm Ltd., Cambridge, UK}
  \country{USA}
}

\author{Shuyue Stella Li}
\email{sli136@jhu.edu}
\affiliation{%
  \institution{Department of Computer Science, Johns Hopkins University}
  \city{Baltimore, MD}
  \country{USA}
}

\author{Andrew N. Sloss}
\email{andrew.sloss@arm.com}
\affiliation{%
  \institution{Arm Ltd., Cambridge, UK}
  \country{USA}
}

\author{Kenneth N. Reid}
\email{ken@kenreid.co.uk}
\affiliation{%
 \institution{Department of Animal Science, Michigan State University}
 \city{East Lansing, MI}
 \country{USA}
  \postcode{48824}
}

\author{Yuan Yuan}
\authornote{Now at Beihang University, Beijing, China}
\email{yyuan@msu.edu}
\affiliation{%
  \institution{Department of Computer Science and Engineering, Michigan State University}
  \city{East Lansing, MI}
  \postcode{48824}
  \country{USA}
}

\author{Wolfgang Banzhaf}
\email{banzhafw@msu.edu}
\affiliation{%
  \institution{Department of Computer Science and Engineering, Michigan State University}
  \city{East Lansing, MI}
  \country{USA}
  \postcode{48824}
}

\renewcommand{\shortauthors}{Peeler and Li, et al.}

\newacro{ACO}[ACO]{Ant Colony Optimization}
\newacro{ACOTSP}[ACOTSP]{Ant Colony Optimization for the Traveling Salesman Problem}
\newacro{EA}[EA]{Evolutionary Algorithm}
\newacro{EC}[EC]{Evolutionary Computation}
\newacro{GP}[GP]{Genetic Programming}
\newacro{LGP}[LGP]{Linear Genetic Programming}
\newacro{OSL}[OSL]{Osaka List Structure}
\newacro{SSP}[SSP]{Subset Sum Problem}


\begin{abstract}
In this paper we introduce Shackleton as a generalized framework enabling the application of linear genetic programming 
to a variety of use cases. We also explore here a novel application for this class of methods: optimizing sequences of LLVM optimization passes. The algorithm underpinning Shackleton is discussed, with an emphasis on the effects of different features unique to the framework when applied to LLVM pass sequences. Combined with analysis of different hyperparameter settings, we report the results on automatically optimizing pass sequences using Shackleton for two software applications at differing complexity levels. Finally, we reflect on the advantages and limitations of our current implementation and lay out a path for further improvements. These improvements aim to surpass hand-crafted solutions with an automatic discovery method for an optimal pass sequence.
\end{abstract}




\maketitle


\section{INTRODUCTION} \label{sec:Introduction}

    \ac{LGP} \cite{bb2007} is categorized within the super-set of \ac{GP} methodologies. These methodologies were established in the early 1990s as a means to generate algorithms and programs automatically, steered by an evolutionary process \cite{koza1992,bnkf1998}. Historically, the proliferation of \ac{LGP} (and the wider umbrella of \acp{EA} it sits under) has been hindered by the need to tailor each implementation to its target application and the limited selection of potential applications \citep{comparisonLGPtech}. As a result, there has been and continues to be an interest in frameworks designed to make \acp{EA} and \ac{LGP} methods more accessible to a wider audience. With such frameworks, experimentation demonstrates how these methods fare in new problem spaces. However, the finer details of their implementation - their coding language, provided examples, overall structure - bring trade-offs that tend toward complexity and difficulty in comprehension \citep{DEAP}. In this paper, we introduce the Shackleton Framework, a generic \ac{LGP} framework that intends to tackle some of these shortcomings.

    Shackleton is a generalized framework that allows for the exploration of applying \ac{LGP} 
    to novel use cases with minimal background knowledge. The aim of the initial creation of Shackleton and this paper is two-fold: (i) to explore the capabilities and usability of a generic \ac{LGP} framework written entirely in the C language, and (ii) to assess the performance of such a framework on the optimization of a practical and complex use-case: the optimization of LLVM compiler optimization pass sequences. The choice of C as the programming language for this project is motivated by its efficiency and its rarity as a language for \ac{EA} methods (explored more in Section \ref{sec:RelatedWork}). Compiler optimization passes are also a rare target for \acp{EA} despite the creation of effective pass sequences being a problem-space exploration problem. Through creation of the framework and hyperparameter experimentation, we explore the implications of these choices and provide an analysis of the optimization sequences that Shackleton's evolutionary scheme generates.
    


     This paper describes the framework in detail, with an emphasis on what distinguishes it from other \ac{LGP} methods. Section \ref{sec:RelatedWork} discusses related work and important background topics to understand. Section \ref{sec:methods} introduces Shackleton in full, outlines the problem space to be explored, and the hyperparameter tuning methodology used. Section \ref{sec:experiments} shares the setup and methodology, results, and discussion on four distinct experiments. Finally, section \ref{sec:conclusion} concludes and summarizes this paper, and outlines potential future applications. \\


    

\section{Background and Related Work} \label{sec:RelatedWork}

    The work presented in this paper is a combination of method, implementation, and target application, but is built upon structures created by other researchers and software developers. 

    Due to the often problem-specific nature of \acp{EA}, a common task explored is the creation of \ac{EA} frameworks that decrease barriers to experimentation. Some example frameworks are generalized, covering many methods. No matter the method, these frameworks offer advantages over writing entire solutions from scratch. One such framework is the Distributed \acp{EA} in Python (DEAP) package \citep{DEAP}, which utilizes the increasingly popular Python language with a wide range of features. There are other frameworks that focus specifically on \ac{GP} but with more novel language choices, such as PushGP \citep{Push} and its derivatives: PyshGP \citep{PyshGP} written in Python, and Clojush \citep{Clojush} written in Clojure. `Push' itself is a language written specifically to support \ac{GP} experimentation and operators. Shackleton shares this attribute with Push, but is written entirely in C to leverage its level of memory control and runtime efficiency. Paired with Shackleton's customizability through the integrated editor tool, our framework is unique amongst GP frameworks as well as being specifically a \textit{linear} \ac{GP} framework.

    While Shackleton is by design a general framework that can be used with multiple target applications, it was designed with a target of compiler optimization sequences in mind. A compiler completes the sole task of translating computer code written in one source language to a target language in an executable form. In contrast, an \textit{optimizing} compiler tries to also minimize or maximize attributes of the outputted computer program in the target language to achieve some benefit at execution time \citep{asuCompilers1986}. For much of the history of optimized compilation, decisions regarding optimization were made based on a priori static information. Recent research aims to improve this by targeting machine learning based compilation, leveraging the exploration capabilities and increase in popularity of machine learning techniques at the compilation stage \citep{wang2018machine}. \ac{GP} and GAs as methods are no exception to this. A genetic algorithm approach has been applied to GNU Compiler Collection (GCC), a compiler similar to LLVM but differentiated by its fixed number of supported languages and reuse constraints \citep{GAapproachBallal, definitiveguideGCC}. Further, there is precedent for applying \ac{GP} specifically to a compiler context to leverage \ac{GP}'s optimization potential for well-known compiler heuristics \cite{CompilerHeuristicStephenson}. Shackleton builds on this motivation by adding another layer of specificity: optimizing a particular stage of the LLVM Compiler Project's core framework.


    \subsection{Linear Genetic Programming} \label{subsec:lineargeneticprogramming}

        Shackleton as a system relies on a method called \ac{LGP} which descends from a long line of computational concepts and techniques. To understand \ac{LGP}, it is important to understand the evolution of methods that resulted in its definition. \ac{LGP} shares a common ancestor with methods like \textit{cellular automata} and \textit{neural computation}, that are \textit{nature-inspired} or \textit{bio-inspired computation}. This term encompasses a wide variety of methods ranging from those that simulate biological natural phenomena in digital forms to methods that actively employ natural materials in their execution. Within the class of methods that takes inspiration from biology and aims to simulate some biological phenomena, we find a subset of methods known as \ac{EC} or \acp{EA} \citep{NaturalComputingRosenburg}.

        \ac{GP} is an ``\ac{EC} technique that automatically solves problems without requiring the user to know or specify the form or structure of the solution in advance'' \citep{FieldGuideGP}. The Field Guide to \ac{GP} offers that ``at the most abstract level GP is a \textit{systematic}, \textit{domain-independent} method for getting computers to solve problems \textit{automatically} starting from a \textit{high-level statement} of what needs to be done''. This compelling idea spawns a myriad of different sub-methods and projects. \ac{LGP} is one such method where programs are linear sequences of instructions, leveraging the fact that computers often represent and run programs in a linear fashion. This same concept is used in Shackleton.

    \subsection{The LLVM Project} \label{subsec:llvmproject}

        The LLVM Project is a collection of modular and reusable compiler and tool-chain technologies. LLVM utilizes its own intermediate representation (LLVM IR) in its core libraries and processes, allowing its compilation process to be largely source- and target-independent \citep{LLVMProject}. During the compilation process, the LLVM IR is optimized using sequences of LLVM passes, optimizations that traverse some portion of the program to either collect information (analysis passes) or transform the program (transform passes). This process and specifically the included passes and their order in the final pass sequence is central to the work discussed in this paper. Creating effective pass sequences typically requires prior knowledge of LLVM. Without this knowledge, the next alternative is to use often sub-optimal default pass sequences. Shackleton can create custom pass sequences for a specific application automatically, without the need for domain-specific or LLVM-specific knowledge.

        For the purposes of this work, Shackleton will not explore the entirety of the LLVM optimization pass search space. Analysis passes that are targeted for debugging or visualization purposes are omitted by design. This omission is to avoid conflicts with automation of the framework. All utility passes are also omitted, due to their small number, their considerable focus on visualization, and odd function compared to analysis and transform passes. Additional passes are omitted on a case-by-case basis depending on observed behavior in experimentation. Custom sequences created by the framework are compared against all the default sequences provided in a base LLVM installation.


\section{Methods} \label{sec:methods}

    \subsection{Problem Space} \label{subsection:methods}

        Shackleton is designed to be applied to various use cases, and here we focus on the LLVM use case as described above. The framework takes in the source code of a target program, and outputs the near-optimal optimization pass sequence that the LLVM compiler can use to optimize this program.
        In our experiments, \ac{ACOTSP} \citep{stutzle2002acotsp} and the Backtrack Algorithm for the \ac{SSP} \citep{SSP} are used as the target programs.

        The \ac{ACO} algorithm is inspired by the foraging behavior of some ant species \citep{dorigo2006ant}, in which the path with more visits would be reinforced after each iteration. In the traveling salesman problem, a graph of cities and the distance between each pair of cities are given, and the goal is to find a path in the topology that minimizes total distance travelled.

        In the \ac{SSP}, we are given a set of $N$ numbers from 1 to 1,000,000, and we are asked to find a subset that sums up to a random number $X$ (where $1,000,000 < X < N \times 1,000,000$) \citep{SSP}\citep{civicioglu2013backtracking}\citep{alekhnovich2011toward}\citep{lagarias1985solving}. Although it might run into halting states, the backtracking algorithm approximates the solution in a reasonable time. Both \ac{ACOTSP} and \ac{SSP} are NP-complete problems, as they do not have polynomial-time solutions. Therefore, optimizing the algorithms from the compiler level is of great importance.

    \subsection{The Shackleton Framework} \label{subsec:shackletonframework}

        Shackleton is a generic \ac{GP} framework that aims to make \ac{GP} easier for a myriad of uses. Currently, the main target of Shackleton is to use the framework for optimization of LLVM pass sequences that ultimately optimize executable code. The source code for Shackleton is publicly available on \href{https://github.com/ARM-software/Shackleton-Framework}{\color{blue}Arm's GitHub page}.

        Shackleton is underpinned by a generic data structure that represents what will be evolved in Shackleton. Any object type can reside in this structure, an attractive attribute given how often problem-specific implementations are needed to achieve favorable results \citep{ProblemSpecificEA}. A unique offering of Shackleton is the editor tool that enables new use cases to be formulated for the tool with ease: this may be explored in more detail in supplemental materials.
        

        \subsubsection{Genetic Programming Design in Shackleton} \label{subsubsec:GPDesigninShackleton}

            \begin{figure}
             \includegraphics[width=\linewidth]{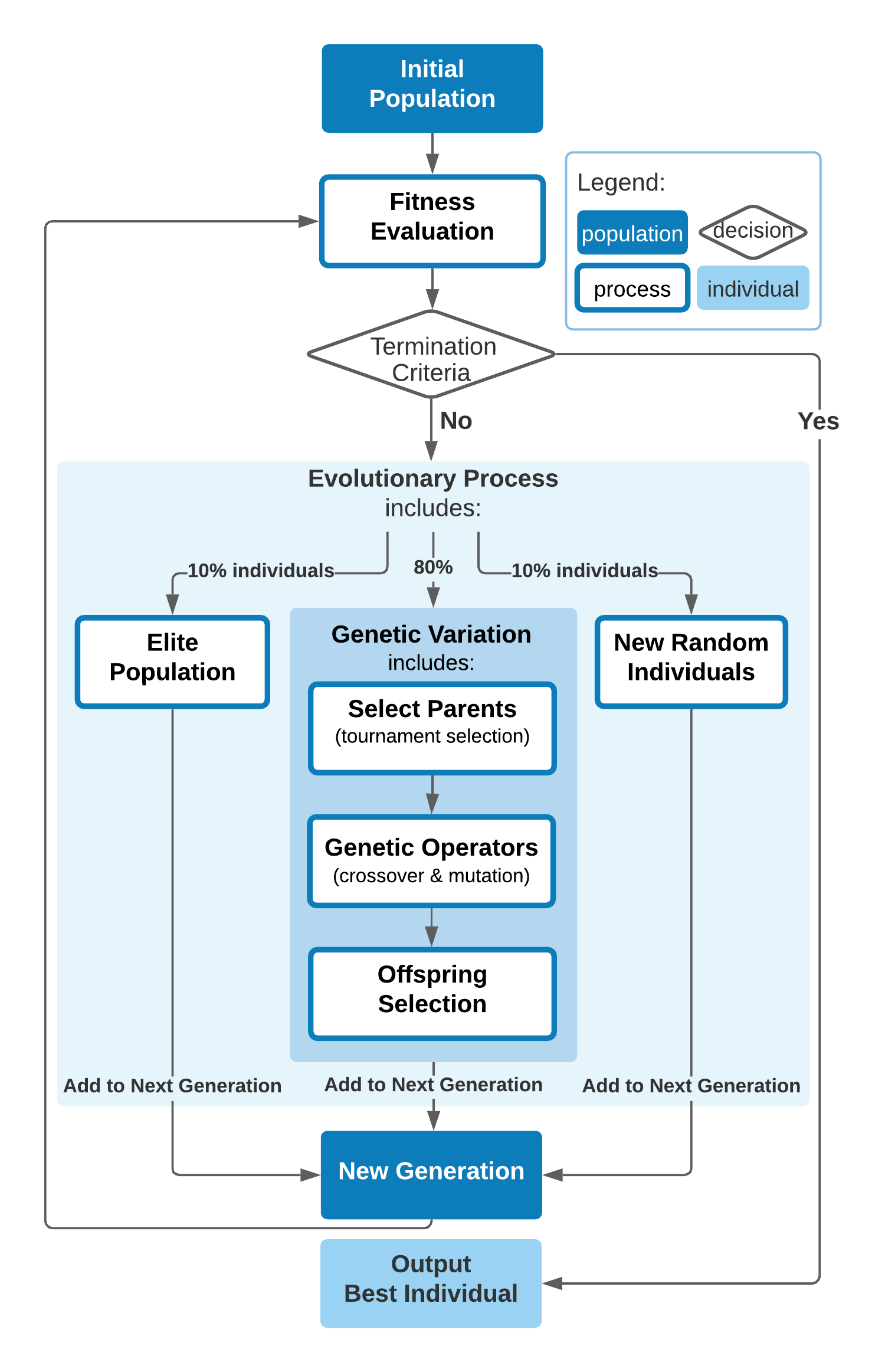}
             \caption{Novel Genetic Programming}
             \label{fig:novelgp}
            \end{figure}
            
            \begin{figure*}[htp]
             \includegraphics[width=.8\linewidth]{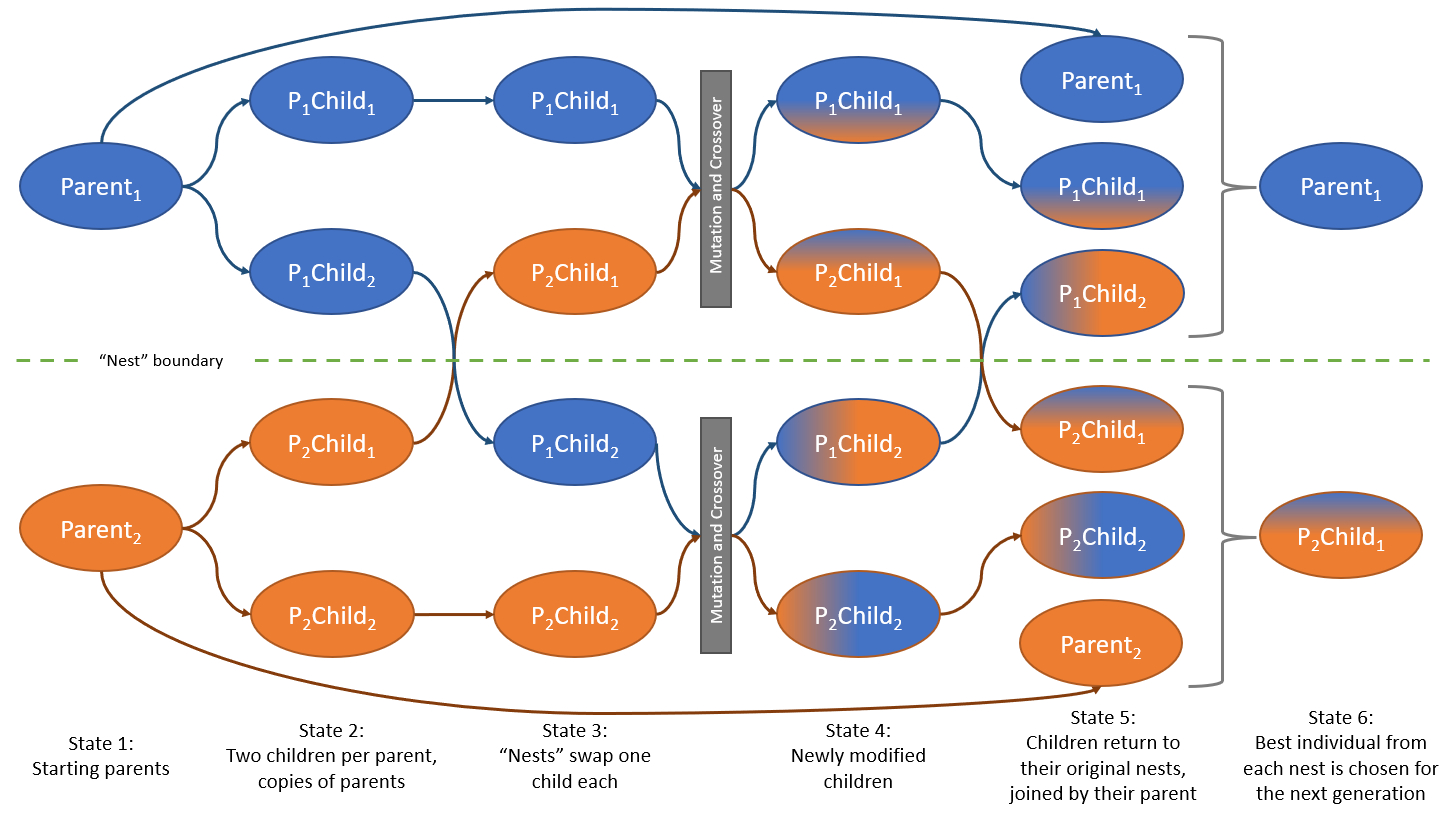}
             \caption{Offspring Selection Process. This does not show a deterministic outcome that will always occur, but rather a possible outcome from a single iteration of the stochastic process. The coloration represents the genetic identity of an individual: the same color means identical genetic materials, and gradients represent a mixing of material from multiple individuals.}
             \label{fig:offspringselect}
            \end{figure*}
            
            We now describe the implementation of \ac{LGP} in the Shackleton Framework in detail, which includes elitism, new individuals, and a special offspring selection mechanism as shown in FIG \ref{fig:novelgp}. In the LLVM flag use case of Shackleton, each chromosome consists of a sequence of optimization flags. First, a population of random individuals is generated. For each sequence, its fitness is calculated by compiling the program with that sequence of passes and averaging the program runtime over 40 runs. In each generation, a portion of the individuals with the best fitness scores are chosen as the elite group and copied over to the next generation unchanged. Then, in the special offspring selection process, a tournament-style selection approach is used to select parent individuals. Genetic operators including one-point crossover and one-point substitution mutation are applied to their genomes to produce a `nest' of 4 offspring individuals; in each nest, the two best performing individuals out of the four offspring and the two parents are selected to join the next generation. This special offspring selection mechanism is inspired by Tacket's brood selection \citep{tacket1994}. Finally, some new individuals are randomly generated to be included in the next generation in order to preserve diversity in the population. The fitness of the new generation is then evaluated and evolution continues. Throughout this process, the genes that would not contribute to faster programs are gradually eliminated from the population, as they are more likely to possess genomes with worse fitness values. The program stops when the termination criteria are met and it outputs individuals with the best fitness scores as the final solution. For the experiments that are run in this paper, the termination criterion is when the number of generations reaches the set limit, but other implementations (e.g., converging fitness values) are also available and can be further explored.

    \subsection{Hyperparameter Tuning} \label{subsec:hyperparammethod}
        
        Shackleton's algorithm requires that a selection of hyperparameters be set ahead of runtime. These hyperparameters affect the structure of the evolutionary process (i.e., number of generations, population size, and tournament size) and the balance of the application of evolutionary operators (i.e., mutation, and crossover). There are additional hyperparameters such as the percentage of elite individuals carried to the next generation, nest size in the special offspring selection mechanism, length of each individual, as well as different fitness metrics. It is often unclear what the optimal combination of hyperparameters is for a given application or implementation.

        In the experimental section of this paper, we first test out extreme combinations of the number of generations and number of individuals to determine the effect of these parameters. Then, all hyperparameter values are fixed at a moderate level to explore three algorithmic designs that consist of different ways to calculate the fitness value of each individual. Lastly, the effects of crossover rate, mutation rate, and tournament size are explored in the hyperparameter tuning process using the Taguchi Method \citep{freddi_salmon_2019}. With the Taguchi Design, as few as 16 sets of parameter combinations are needed to explore 4 levels of the three parameters, which would require 64 experiments in a regular factorial design. We chose these specific hyperparameters because they were found, through preliminary testing, to influence the optimization significantly.

        The experiments were conducted on HPCC nodes running CentOS Linux version 7 and Clang version 8.0.0. We utilized the Backtrack Algorithm (from \ac{SSP}), which contains recursive loops that are potential sites of optimization. Preliminary testing found this problem to be of adequate complexity for our experimentation purposes, considering both the runtime and human-verifiable testability.

\section{Experiments} \label{sec:experiments}

    \subsection{Number of Generations vs. Population Size}\label{subsection:gensVSpopSize}
        
        \subsubsection{Experiment Setup}
            
            The number of generations and population size directly determine the number of runtime evaluations. These two hyperparameters are roughly proportional to how many individuals will appear throughout the evolutionary process. Therefore, in order to reduce the overhead runtime of Shackleton, it is essential to determine how much these two factors influence the search process of the optimal solution. In our first experiment, we tracked the fitness of the best individual in each generation with extreme settings of number of generations and population size. This is to see whether more computing power and runtime should be allocated to a large number of generations or a larger population size, when their product is fixed. The values are set such that each parameter setting has a constant product, so the same number of total individuals are represented in the entire run of Shackleton. 

            Table \ref{tab:Extreme Parameter Experiment} shows the combinations of parameters used for this experiment. Six trials were run for each setting and other hyperparameters are kept constant.
    
            \begin{table}
            \caption{\label{tab:Extreme Parameter Experiment} \textbf{Number of Generations vs. Population Size Experiment Parameters}}
            \label{tab:extreme}
            \begin{tabular}{ccc}
            \toprule
            Experiment Nr. & Number of Generations & Population Size \\
            \midrule
             0 & 50 & 40 \\
             1 & 250 & 8 \\
             2 & 200 & 10 \\
             3 & 10 & 200 \\
             4 & 8 & 250 \\
             5 & 4 & 500 \\
            \bottomrule
            \end{tabular}
            \end{table} 

        \subsubsection{Results}
        
\begin{figure}
                \includegraphics[width=\linewidth]{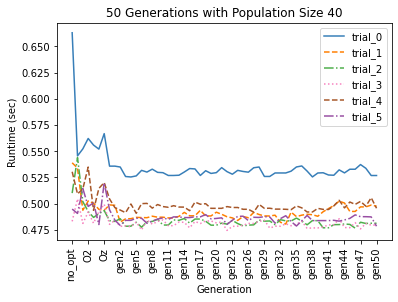}
                \caption{gen=50, pop=40}
                \label{fig:extremeparamscontrol}
            \end{figure}

\begin{figure}
                \includegraphics[width=\linewidth]{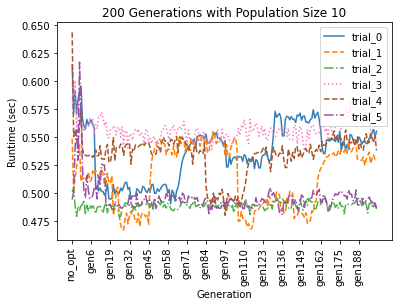}
                \caption{gen=250, pop=8}
                \label{fig:extremeparamslargegen}
            \end{figure}

\begin{figure}
                \includegraphics[width=\linewidth]{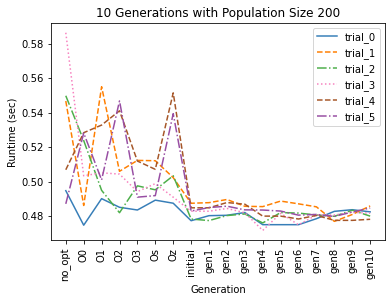}
                \caption{gen=10, pop=200}
                \label{fig:extremeparamslargepop}
            \end{figure}

\begin{figure}
                \includegraphics[width=\linewidth]{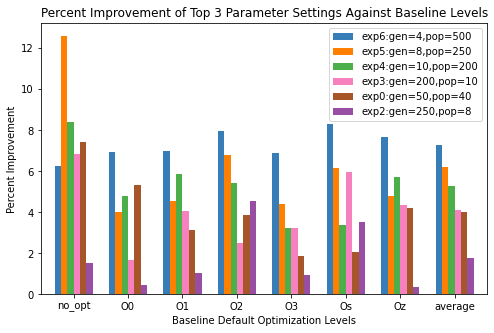}
                \caption{Improvement over Baseline}
                \label{fig:extremeparamsbar}
            \end{figure}

            FIG \ref{fig:extremeparamscontrol} plots the control set up with $50$ generations and a population size of $40$; FIG \ref{fig:extremeparamslargegen} and FIG \ref{fig:extremeparamslargepop} plot two representative cases of a setting with a large \textit{num\_generation} value and a setting with a large \textit{population\_size} value. The $y$-axis is the runtime of the sample use case - the \ac{SSP} with the Backtrack Algorithm - in seconds, which we use as the fitness of the individuals, and the $x$-axis is the generation number. The runtime of the program with no optimization and with default LLVM optimization flags (first 8 data points of each trend line) is plotted with the runtime of the individuals across generations. FIG \ref{fig:extremeparamsbar} shows the percentage improvement of the parameter settings compared to each of the baseline optimization levels. 
            
        \subsubsection{Discussion}
            
            By exploring the extreme ends of the search space, this experiment shows that more computing power and runtime should be allocated to the population size rather than the number of generations. In the hyperparameter setting plotted in FIG \ref{fig:extremeparamslargegen}, a small population size of $10$ is allowed to undergo $200$ generations in search for the optimal solution, but the fitness of the best individual is noisy and does not tend to converge, with a final percent improvement of 4.08\%. On the other hand, a large population size of $200$ shows converging fitness within just $10$ generations and a final percent improvement of 5.25\%. This result gives support to the claim that the greater diversity in a large population is crucial in the search for an optimal solution.

    \subsection{Fitness Calculation}
        
        \subsubsection{Experiment Setup}
        
        The framework is implemented with three algorithmic designs that feature different ways of calculating the fitness scores, as well as re-evaluating repeated individuals. In calculating the runtime of each individual, the optimization sequence is applied to the source code and the runtime of the code is taken as the average across 40 runs. Observations from preliminary testing indicate that runtime across different runs can vary considerably. Therefore, we first hypothesize that using the sum of the average runtime and the variance as the fitness might prefer individuals with more consistent runtime and lead to better optimization. We also hypothesize that for the same optimization sequence, storing the runtime into a hash table instead of re-doing the 40 runs each time an individual appears in the population could decrease overall runtime. This section explores the impacts of these two potential improvements in terms of algorithmic design. 

        For each fitness calculation, all hyperparameters are kept at a moderate level as shown in Table \ref{tab:defaultparam}. Table \ref{tab:fitnesscal} highlights the different designs.

        \begin{table}
        \caption{\textbf{Fitness Calculation Experiment Setup}}
        \label{tab:fitnesscal}
        \begin{tabular}{cccc}
        \hline
         Experiment Nr.  &  Fitness  &  Hashing  &  \% Improvement  \\
        \hline
         6 & runtime & Yes & 11.29 \\
         7 & time+var & Yes & 1.27 \\
         8 & runtime & No & 3.09 \\
        \hline
        \end{tabular}
        \end{table}
        
        \begin{table}
        \caption{\textbf{Default Hyperparameter Settings}}
        \label{tab:defaultparam} 
        \begin{tabular}{lc}
         \toprule
         Hyperparameter & Value \\
         \midrule
         number of generations & 50 \\
         population size & 40 \\
         crossover probability & 50 \\
         mutation probability & 50 \\
         tournament size & 4 \\
         elite percentage & 10 \\
         nest size & 6 \\
        \bottomrule
        \end{tabular}
        \end{table}
        
        \subsubsection{Results}
        
\begin{figure}
             \includegraphics[width=\linewidth]{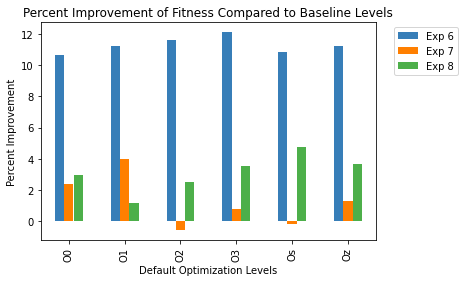}
             \caption{Percent Improvement of Different Fitness Calculations}
             \label{fig:algosfitnesscmp}
            \end{figure}

\begin{figure}
             \includegraphics[width=\linewidth]{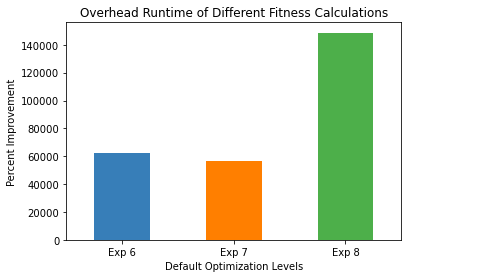}
             \caption{Overhead Runtime of Different Fitness Calculations}
             \label{fig:algosoverheadcmp}
            \end{figure}

            The fitness across generations and the overhead runtime of Shackleton are recorded as the key metrics of this experiment. FIG \ref{fig:algosfitnesscmp} shows the percent improvement of fitness compared to the default LLVM optimization levels. Fitness evaluation using only the runtime shows a significant advantage compared to 
            using both the runtime and the variance across runs. FIG \ref{fig:algosoverheadcmp} shows the overhead runtime of each design taken as the average across six trials. 
            
        \subsubsection{Discussion}
            
            We compared experiment \verb|#|6 and experiment \verb|#|7, which differ only in that the fitness includes the variance of the runs in experiment \verb|#|7. Experiment \verb|#|6 is able to achieve a larger percentage improvement. This implies that the variance amongst runs of a program is not relevant to the optimization sequence applied to it, but rather due to the inherent fluctuations of clock speed caused by sharing of resources on the same computing cluster. Therefore, including the variance into the fitness function would not help reduce the noise in the data and could distract the population from moving toward a sequence with shorter runtime. 

            For experiment \verb|#|8, where the fitness is re-evaluated and averaged every time an individual appears in the population, the more comprehensive fitness calculation greatly increases Shackleton's overhead runtime. However, as shown by the percent improvement after the genetic algorithm, this did not lead to a more effective evolution process. Therefore, the improvement of looking up the fitness value from a hash table helps considerably with the overall runtime and does not sacrifice performance
            . In future experiments, the fitness will be calculated the same way as it is in experiment \#6.

    \subsection{Hyperparameter Tuning}
        
        \subsubsection{Experiment Setup}
        
        In this experiment, the optimal hyperparameter combination for percent crossover, percent mutation, and tournament size is explored using a Taguchi design \citep{freddi_salmon_2019}. Other hyperparameters are kept constant: 50 generations, population size = 40, elite percentage = 15, nest size = 6, and individual size = 0 (When individual size is set to $0$, individuals will be randomly generated with length between 10 and 90. Single-point mutation keeps the length constant, and crossover changes the length of the offspring). The three independent hyperparameters are tested across 4 equidistant levels; the Taguchi Method effectively reduces the number of combinations needed to 16 compared to $4^3=64$ in a regular factorial grid search design. 
        
        \begin{table}
        \caption{\textbf{Parameter Tuning Experiment Setup}}
        \label{tab:paramtuning} 
        \begin{tabular}{ccccc}
         \toprule
        Exp. & Crossover & Mutation & Tournament & Improvement\\
        Nr. & & & Size & \\
        \midrule
         09 & 20 & 20 & 2 & 5.44\%\\
        
         10 & 20 & 40 & 4 & 0.50\% \\
        
         11 & 20 & 60 & 6 & 7.04\% \\
        
         12 & 20 & 80 & 8 & 1.91\% \\
        
         13 & 40 & 20 & 4 & 4.88\% \\
        
         14 & 40 & 40 & 2 & 3.49\% \\
        
         15 & 40 & 60 & 8 & 5.76\% \\
        
         16 & 40 & 80 & 6 & 4.90\% \\
        
         17 & 60 & 20 & 6 & 8.31\% \\
        
         18 & 60 & 40 & 8 & 4.18\% \\
        
         19 & 60 & 60 & 2 & 3.43\% \\
        
         20 & 60 & 80 & 4 & 10.88\% \\
        
         21 & 80 & 20 & 8 & 5.62\% \\
        
         22 & 80 & 40 & 6 & 5.53\% \\
        
         23 & 80 & 60 & 4 & 7.21\% \\
        
         24 & 80 & 80 & 2 & 6.83\% \\
         \bottomrule
        \end{tabular}
        \end{table}
        
        \subsubsection{Results}
        
\begin{figure}
         \includegraphics[width=\linewidth]{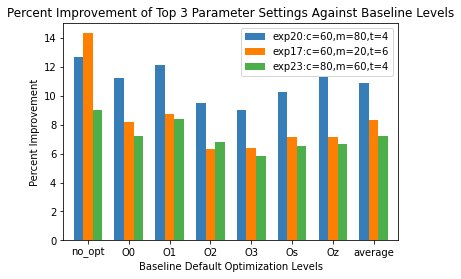}
         \caption{Top 3 Hyperparameter Combinations}
         \label{fig:top3paramimprovA}
        \end{figure}

\begin{figure}
         \includegraphics[width=\linewidth]{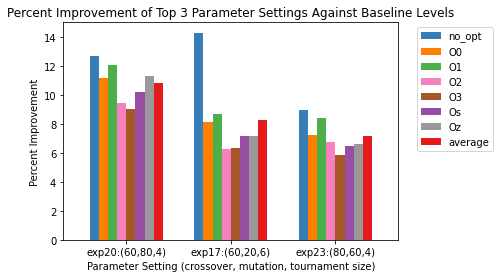}
         \caption{Top 3 Hyperparameter Combinations}
         \label{fig:top3paramimprovB}
        \end{figure}

\begin{figure}
         \includegraphics[width=\linewidth]{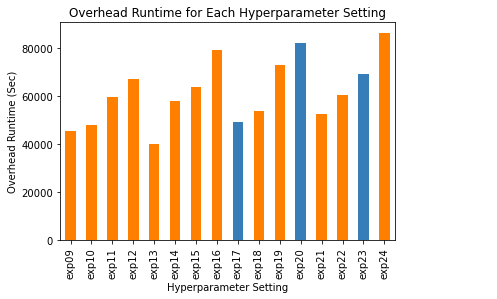}
         \caption{Overhead Runtime of Shackleton Framework}
         \label{fig:overheadruntime}
        \end{figure}

        Table \ref{tab:tophyperparameters} shows the three hyperparameter combinations that yield the highest percent improvement in terms of runtime when compared against each default LLVM optimization level. From Table \ref{tab:paramtuning}, it can be concluded that the combination (60,80,4) from experiment $20$ - crossover probability 60\%, mutation probability 80\%, and tournament size of $4$ - is the optimal hyperparameter combination as it is consistently the top-performing combination compared against any default optimization level. Examining the top 3 hyperparameter combinations, we can see that (60,20,6) and (80,60,4) also give consistently good optimization. FIG \ref{fig:top3paramimprovA} and \ref{fig:top3paramimprovB} shows the percent improvement in terms of runtime compared with the baseline optimizations levels using the above three best hyperparameter combinations. The overhead runtime averaged over six trials for each parameter setting is plotted in FIG \ref{fig:overheadruntime}, with the top three performing hyperparameter combinations in blue and others in orange. Lastly, a Taguchi Analysis \citep{freddi_salmon_2019} is performed to analyze the importance factor of the three hyperparameters, and the results are shown in Table \ref{tab:taguchitable}.
        
        \begin{table}
        \caption{\textbf{Top Hyperparameter Combinations}}
        \label{tab:tophyperparameters} 
        \begin{tabular}{cccc}
         \toprule
         Baseline & Top 1 & Top 2 & Top 3 \\
         \midrule
         No opt & (60,80,4) & (60,20,6) & (80,60,4) \\
         Improvement & 19.12\% & 18.20\% & 10.43\% \\
         \hline
         
         O0 & (60,80,4) & (40,20,4) & (60,20,6) \\
         Improvement & 17.17\% & 11.13\% & 9.17\% \\
         \hline
         
         O1 & (60,80,4) & (20,60,6) & (80,60,4) \\
         Improvement & 18.31\% & 10.83\% & 10.46\% \\
         \hline
         
         O2 & (60,80,4) & (80,60,4) & (80,80,2) \\
         Improvement & 15.27\% & 8.31\% & 7.88\% \\
         \hline
         
         O3 & (60,80,4) & (20,60,6) & (60,20,6) \\
         Improvement & 14.53\% & 8.24\% & 7.18\% \\
         \hline
         
         Os & (60,80,4) & (80,80,2) & (60,20,6) \\
         Improvement & 16.01\% & 8.22\% & 8.14\% \\
         \hline
         
         Oz & (60,80,4) & (60,20,6) & (80,60,4) \\
         Improvement & 18.83\% & 8.19\% & 8.00\% \\
         \hline
         
         Average & (60,80,4) & (60,20,6) & (80,60,4) \\
         Improvement & 17.04\% & 9.69\% & 8.57\% \\
         \bottomrule
        \end{tabular}
        \end{table}
        
        \begin{table}
        \caption{\textbf{Taguchi Analysis - Improvement Signal-to-Noise Ratio}}
        \label{tab:taguchitable} 
        \begin{tabular}{cccc}
        \toprule
         & Crossover & Mutation & Tournament Size\\
        \midrule
        L1 & -7.92 & 2.14 & 7.09 \\
        L2 & 3.26 & -4.94 & -13.00\\
        L3 & 5.01 & 7.44 & 4.08 \\
        L4 & 3.87 & -0.41 & 6.06 \\
        $\Delta$ & 12.93 & 12.38 & 20.10 \\
        \bottomrule
        \end{tabular}
        \end{table}
        
        \begin{table}
        \caption{\textbf{Taguchi Analysis - Overhead Runtime Signal-to-Noise Ratio} }
        \label{tab:taguchitabletime} 
        \begin{tabular}{cccc}
        \toprule
         & Crossover & Mutation & Tournament Size\\
        \midrule
        L1 & 48.84 & 49.71 & 44.25 \\
        L2 & 46.63 & 47.53 & 48.60\\
        L3 & 43.39 & 43.78 & 42.20 \\
        L4 & 43.88 & 41.71 & 47.68 \\
        $\Delta$ & 5.45 & 8.00 & 6.40 \\
        \bottomrule
        \end{tabular}
        \end{table}
        
        \subsubsection{Discussion}
        
        From Table \ref{tab:tophyperparameters}, it can be observed that the top hyperparameter configuration when compared to any baseline optimization level is when crossover probability equals 60, mutation probability equals 80, and tournament size equals 4. Most hyperparameter combinations that made it to the top 3 when compared to any baseline level have a high rate of crossover, high rate of mutation, and a moderate tournament size. Parents from the previous generation are selected in a tournament style, where $n$ individuals will be randomly chosen from the entire population, and the individual with the best fitness is taken as the origin of a genetic operator. A higher tournament size will result in the parent being selected to have a higher fitness, but setting it too high will reduce population diversity as it results in the same individuals being selected every time. Crossover and mutation are genetic operators that make changes from the original parents. High crossover and mutation probabilities contributes to the diversity of the offspring population. Therefore, a moderate tournament size ensures that `good enough' individuals are selected as parents. High crossover and mutation probabilities maintain diversity in the population so that the search can keep moving towards the optimal solution.

        From FIG \ref{fig:overheadruntime}, the overhead runtime roughly correlates to the level of the mutation probability. However, the Taguchi Analysis \citep{freddi_salmon_2019} in Table \ref{tab:taguchitable} shows that tournament size has the largest effect on the optimization process. A larger range means that a change of this hyperparameter can cause a larger effect on the measurement output. Table \ref{tab:taguchitabletime} shows the same analysis done with respect to the overhead runtime of the Shackleton Framework. The range of the signal-to-noise ratio for each hyperparameter is fairly similar, implying that each hyperparameter has a similar level of impact on the overhead runtime. The above information should also be taken into consideration when choosing the best hyperparameter combination.

        Looking at both the performance and runtime metrics and the results from the Taguchi Analysis, experiment \verb|#|17 (crossover probability = 60, mutation probability = 20, and tournament size = 6) is among the top 3 hyperparameter combinations with highest percent improvement and has a reasonable runtime. The high crossover probability maintains the diversity in the population; low mutation probability contributes to the fast overhead runtime; and a large tournament size selects efficient parents without reducing diversity. 

    \subsection{Robustness Analysis}
        
        \subsubsection{Experiment Setup}
        
            Even though the hyperparameter settings are problem-specific and can be set by the user, it is important to make sure that most (if not all) hyperparameter combinations will speed up the execution of the target source code. In this experiment, each of the 16 hyperparameter combinations from Table \ref{tab:paramtuning} is run 30 times, and the data from each of these $16\times 30=480$ runs are collected to examine the robustness of Shackleton. 
        
        \subsubsection{Results}
        
            Two metrics are of importance in this evaluation - percent optimization from the raw source code and the percent improvement of the automatically-generated sequence compared to the default optimization. We first want to ensure that Shackleton will produce optimization sequences that speed up the execution of the target source code. Then, we want the automatically-generated sequences to perform better than the default LLVM optimization.

\begin{figure}
             \includegraphics[width=\linewidth]{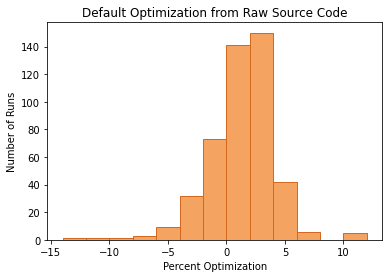}
             \caption{Percent Optimization of All Hyperparameter Settings (compared to raw source code)}
             \label{fig:allo2opti}
            \end{figure}

\begin{figure}
             \includegraphics[width=\linewidth]{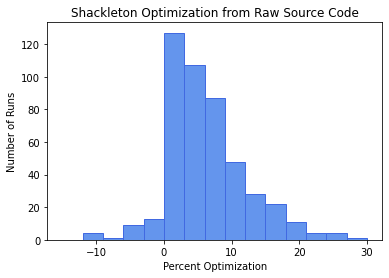}
             \caption{Percent Optimization of All Hyperparameter Settings (compared to raw source code)}
             \label{fig:allparamopti}
            \end{figure}

\begin{figure}
             \includegraphics[width=\linewidth]{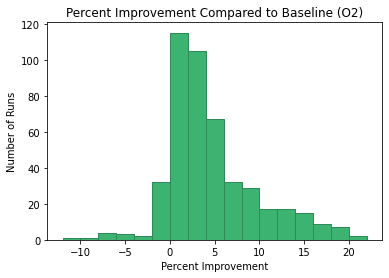}
             \caption{Percent Improvement of All Hyperparameter Settings (compared to O2)}
             \label{fig:allparamimprov}
            \end{figure}

            In FIG \ref{fig:allo2opti}, the percent optimization produced by the default LLVM -O2 sequence is plotted; $74\%$ of these runs produce faster code compared to the unoptimized source code. The histogram in FIG \ref{fig:allparamopti} shows the percent optimization produced by the Shackleton-generated sequence; $94\%$ of these runs produce faster code compared to the unoptimized source code. Lastly, the histogram in FIG \ref{fig:allparamimprov} shows the percent improvement compared to the default -O2 LLVM compiler flag; $90\%$ of the Shackleton-generated optimization runs are faster than the default optimization runs. Outlier values (more than 3 standard deviations from the mean) are removed from the analysis because these data points are not taken as an average among trials, making it more prone to system fluctuations. The experimental results are summarized in Table \ref{tab:robustnessresults}.
            
           \begin{table}
            \caption{\textbf{Percent Improvement}}
            \label{tab:robustnessresults} 
            \begin{tabular}{ccc}
            \toprule
            Comparison & Average \% Improvement & Additional \%\\
            \midrule
            O2-raw & 1.40 & 74.09 \\
            Shackleton-raw & 6.20 & 93.88 \\
            Shackleton-O2 & 4.77 & 90.17 \\
            \bottomrule
            \end{tabular}
            \end{table}
            
        \subsubsection{Discussion}
            
            
            Shackleton's use of stochastic elements in its evolutionary scheme mean that speed improvements over the raw source code after optimization are not always guaranteed. However, the experiments shown here indicate that Shackleton consistently produces optimized code that is faster and offers a larger average improvement than that produced by the default LLVM optimization sequences. These experiments are not exhaustive and notably omit some hyperparameters. More comprehensive analysis of these omitted hyperparameters and additional scaling of the parameters shown here could offer additional insights into benefits and limiting factors. Based on our experiments, with fine-tuned hyperparameters the optimization pass sequences automatically generated by Shackleton can be expected to achieve even better performance.

\section{CONCLUSION AND FUTURE WORK} \label{sec:conclusion}
    
    The existing predefined LLVM optimization levels (-Ox) are used for general purpose program optimization and are not problem specific. Targeted runtime optimization of a program by a hand-crafted LLVM pass or a pass sequence would require expert knowledge in both LLVM and the program to be optimized. In this paper, we presented the Shackleton Framework, which is able to automatically generate near-optimal LLVM optimization flag sequences. The Shackleton Framework, without any prior knowledge of the compiler, the optimization passes, nor the program, consistently produces optimization pass sequences that achieve significant runtime improvement over 
    default optimization options. We first proposed an elitist approach and a special offspring selection mechanism for a more greedy genetic algorithm, while a new random offspring mechanism ensures diversity in the population. We also performed hyperparameter tuning and explored the effect of each hyperparameter on the optimization results and overhead runtime. 
    
    There are numerous opportunities to further improve and analyze the Shackleton Framework. Shackleton currently only supports a relatively small subset of LLVM optimization passes and only compares against the default optimization passes for analysis. Widening the scope of passes included and allowing for comparison against bespoke sequences fit for a specific use case could offer benefits beyond what is shown here. LLVM also allows programmers to create their own passes to leverage attributes of their specific applications more effectively than the base set of passes. This benefit of the open-source framework could be integrated into Shackleton to generate better sequences faster. In addition to benefits for the LLVM use case, the generic Shackleton Framework could be improved by allowing for hyperparameter tuning beyond what is currently supported. Notably, the brood/nest size and selection percentage for the special offspring selection mechanism are currently fixed. Allowing these parameters to vary or be configured by hand would offer better control over the evolutionary process and may provide more refined results. As an open source framework, Shackleton is designed to be experimented with and expanded upon. With further improvements, the pass sequences automatically generated by the Shackleton Framework could potentially surpass hand-crafted pass sequences while simultaneously freeing developers' time to pursue other advancements for compilers.


\begin{acks}
    This work was generously funded by the EnSURE (Engineering Summer Undergraduate Research Experience) program at Michigan State University (MSU) as well as the John R. Koza Endowment. We also gratefully acknowledge The Institute of Cyber-Enabled Research (ICER) at MSU for providing the hardware infrastructure that made the computation required to complete this work possible. 
\end{acks}

\bibliographystyle{ACM-Reference-Format}
\bibliography{references}

\end{document}